\documentclass[letterpaper, 10pt, conference]{ieeeconf}      

\IEEEoverridecommandlockouts                              

\usepackage{amsmath,amssymb,amsfonts}
\usepackage{enumerate}
\usepackage{lettrine}
\usepackage{graphicx}
\usepackage{textcomp}
\usepackage{makecell}
\usepackage{graphics} 
\usepackage{epsfig}
\usepackage{subfigure}
\usepackage[printonlyused,withpage]{acronym}

\usepackage{epstopdf}
\usepackage{lipsum}

\usepackage{lipsum,amsmath,multicol}
\usepackage{float}
\usepackage{algorithm}
\usepackage{algpseudocode}
\usepackage{wrapfig}
\usepackage{sidecap}

\usepackage{cite}
\usepackage{hyperref}

\usepackage{amsthm}

\usepackage{comment}
\usepackage{array}
\usepackage{glossaries}
\usepackage{breqn}
\makeglossaries
\newcolumntype{L}[1]{>{\raggedright\let\newline\\\arraybackslash\hspace{0pt}}m{#1}}
\newcolumntype{C}[1]{>{\centering\let\newline\\\arraybackslash\hspace{0pt}}m{#1}}
\newcolumntype{R}[1]{>{\raggedleft\let\newline\\\arraybackslash\hspace{0pt}}m{#1}}
\usepackage{sidecap}
\usepackage{url}
\usepackage[T1]{fontenc}
\usepackage{footnote}
\makesavenoteenv{algorithm}
\usepackage{soul}

\usepackage{multicol, blindtext}
\usepackage{multirow}
\usepackage{xcolor}
\usepackage{graphicx}
\usepackage{breqn}

\allowdisplaybreaks

\definecolor{grey}{rgb}{0.5,0.5,0.5}
\newcommand{\diag}{\text{diag}}
\newcommand{\polardir}{\boldsymbol{\omega}}

\makeatother
\begin{document} 

\title{ 
AMSwarm: An Alternating Minimization Approach for Safe Motion Planning of Quadrotor Swarms in Cluttered Environments
}
\author{Vivek K. Adajania, Siqi Zhou, Arun Kumar Singh, and Angela P. Schoellig\thanks{Vivek K. Adajania, Siqi Zhou, and Angela P.~Schoellig are with the Learning Systems and Robotics Lab (http://www.learnsyslab.org) at the University of Toronto Institute for Aerospace Studies, Canada, and the Technical University of Munich, Germany. They are also with the Vector Institute for Artificial Intelligence. Arun Kumar Singh is with the University of Tartu, Estonia. This research was in part supported by the European Social Fund via the ICT program measure and grant PSG753 from the Estonian Research Council.
Emails:
\{vivek.adajania,~siqi.zhou\}@robotics.utias.utoronto.ca,  arun.singh@ut.ee, and angela.schoellig@tum.de.  } 
}
\maketitle

\begin{abstract}
This paper presents a scalable online algorithm to generate safe and kinematically feasible trajectories for quadrotor swarms. Existing approaches rely on linearizing Euclidean distance-based collision constraints and on axis-wise decoupling of kinematic constraints to reduce the trajectory optimization problem for each quadrotor to a quadratic program (QP). This conservative approximation often fails to find a solution in cluttered environments. We present a novel alternative that handles collision constraints without linearization and kinematic constraints in their quadratic form while still retaining the QP form. We achieve this by reformulating the constraints in a polar form and applying an Alternating Minimization algorithm to the resulting problem. Through extensive simulation results, we demonstrate that, as compared to Sequential Convex Programming (SCP) baselines, our approach achieves on average a 72\% improvement in success rate, a 36\% reduction in mission time, and a 42 times faster per-agent computation time. We also show that collision constraints derived from discrete-time barrier functions (BF) can be incorporated, leading to different safety behaviours without significant computational overhead. Moreover, our optimizer outperforms the state-of-the-art optimal control solver ACADO in handling BF constraints with a 31 times faster per-agent computation time and a 44\% reduction in mission time on average. We experimentally validated our approach on a Crazyflie quadrotor swarm of up to 12 quadrotors. The code with supplementary material and video are released for reference.
\end{abstract}

\section{Introduction}

Quadrotor swarms have a great potential in applications such as search and rescue \cite{search_and_rescue}, mapping and environmental monitoring \cite{mapping}, and payload transport \cite{sarah_cable}. As compared to single quadrotors, quadrotors swarms offer increased flexibility, efficiency, and robustness \cite{chung2018survey}.

In this paper, we consider the problem of motion planning for quadrotor swarms in cluttered environments and treat it as a trajectory optimization problem to be solved. In this context, the most straightforward approach is to formulate one joint optimization problem that computes trajectories for all quadrotors. Existing works have used both global mixed-integer linear programming \cite{mip_how} and local optimization-based Sequential Convex Programming (SCP) \cite{augugliaro2012generation} approaches for solving the joint trajectory optimization problem. The solution space of these approaches is large but they quickly become intractable as the number of quadrotors grows. 

Distributed approaches provide a more scalable alternative, where each quadrotor solves an independent trajectory optimization problem taking into account the predicted trajectories of its neighbours \cite{luis-ral19, luis-ral20, dmpc_epfl, soria2021predictive}. The predicted trajectories are often assumed to be shared by the neighbouring quadrotors. These approaches have successfully demonstrated swarm motion planning for tens of quadrotors. However, we will show that their performance in terms of scalability, mission time, and computation time remains poor in highly cluttered environments. While there exist works that further incorporate a high-level discrete planner to improve the swarm performance in cluttered environments~\cite{zhou2017fast,ego_swarm,park2022online}, in this work, we focus on the low-level trajectory optimization problem and propose an algorithm that addresses the limitations of existing distributed trajectory optimization baselines.

\begin{figure}[!t]
    \centering
    \includegraphics[width=\columnwidth]{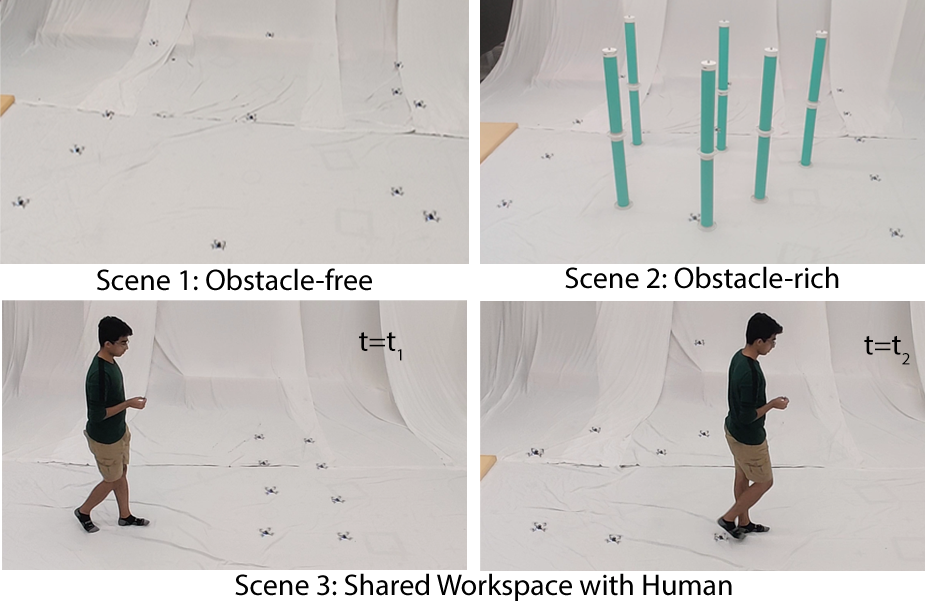}
    \caption{Experimental demonstration of our distributed alternating minimization based approach for quadrotor swarm motion planning in challenging scenes. Link to video: \protect\hyperlink{http://tiny.cc/AMSwarmVideo}{\url{http://tiny.cc/AMSwarmVideo}}. Link to code and supplementary material: \protect\hyperlink{https://github.com/vivek-uka/AMSwarm}{\url{https://github.com/utiasDSL/AMSwarm}}.
    }
    \label{teaser}
\end{figure}

Some of the limitations of  \cite{luis-ral19, luis-ral20,dmpc_epfl} and other related works such as \cite{augugliaro2012generation} and \cite{morgan2014probabilistic} can be attributed to the underlying trajectory optimizer that relies on axis-wise decoupling of kinematic constraints and linearization of collision avoidance constraints. These affine approximations are made to obtain a quadratic program (QP) that can be solved efficiently. However, the computational benefits come at the expense of a reduced solution space. Moreover, while replanning in a receding horizon setting, the collision constraints are not active until the planned trajectories intersect with the neighbouring quadrotors or obstacles \cite{zeng2021safety}. This reduces the responsiveness of the collision avoidance behaviour. One way to mitigate this issue is to use a longer prediction horizon; however, this increases the computation time. 

Our proposed optimizer addresses both limitations discussed above: the conservativeness of existing approaches due to approximations and the late responsiveness to neighbouring quadrotors or obstacles. We show that by reformulating the quadratic kinematic constraints and collision constraints into polar form and applying an Alternating Minimization (AM) algorithm to the resulting optimization problem, we can retain a QP without requiring any linearization. As a result, we obtain more aggressive motions with improved metrics for swarm planning, such as mission time and success rate. Our formulation naturally extends to the case when collision avoidance is modelled by a discrete-time barrier function (BF)~\cite{zeng2021safety}. This dramatically improves safety metrics such as clearance to neighbouring quadrotors and obstacles while incurring no significant computational cost. To the best of our knowledge, only a few works, such as \cite{mali2021incorporating}, \cite{varun2021motion}, have incorporated BF constraints over the entire planning horizon; most works such as \cite{chen2020guaranteed} and \cite{wang2017safety} consider a one-step reactive planning approach. Among the multi-step approaches, ours is the first to formulate trajectory planning with BF constraints as a QP (see Section \ref{gaps_existing}).

We compare our approach with the SCP baselines from \cite{luis-ral19, luis-ral20,dmpc_epfl} and show on average a $72\%$ improvement in success rate, a $36\%$ reduction in mission time, and a $42$ times faster per-agent computation time. Additionally, we show that the proposed approach with BF constraints allows us to introduce different safety behaviours. We further show that our optimizer's handling of discrete-time BF constraints outperforms the state-of-the-art solver ACADO \cite{acado} with a $31$ times faster per-agent computation time and a $44\%$ reduction in mission time on average. 
\raggedbottom
\section{Distributed Motion Planning Problem}
Our goal is to generate smooth, collision-free, and kinematically feasible trajectories that navigate $N$ quadrotors from their initial positions $\mathbf{p}_{i,o}$ to their desired goal positions $\mathbf{p}_{i,g}$ in an obstacle-rich and possibly dynamic environment. The vector $\mathbf{p}=[x,\:y,\:z]^T$ is the three-dimensional position of the quadrotor, the subscript $i$ is the quadrotor index, and the subscripts $o$ and $g$ denote initial and goal variables.

Similar to~\cite{luis-ral19,luis-ral20}, we formulate the quadrotor swarm motion planning as a distributed trajectory optimization problem, where the computation for each quadrotor is parallelized. At each time step, the quadrotors exchange the planned trajectories from the previous step and re-optimize the trajectories towards their goal positions subject to constraints. The distributed optimization problem is solved in a receding horizon fashion until each quadrotor reaches its goal position. 

 We note that, in this work, the quadrotors' trajectories are optimized online to account for possible dynamic obstacles. We assume that the obstacles' current positions and velocities are available to each quadrotor.

\subsection{Optimization Problem Formulation}
\label{subsec:opt_prob_formulation}
 At each planning step, the optimization problem solved by quadrotor $i$ is formulated as follows:

\begin{subequations}
\begin{align}
\min_{\mathbf{p}_i}\hspace{0.5em}& w_{g}\sum_{k=K-\kappa}^{K-1} \left\Vert \mathbf{p}_i[k] - \mathbf{p}_{i,g}\right\Vert^2+w_{s}\sum_{k=0}^{K-1} \left\Vert \mathbf{p}_i^{(q)}[k]\right\Vert^2\label{cost}\\
\text{s.t.}\hspace{0.5em}&     \mathbf{p}_i^{(q)}[0] = \mathbf{p}_{i,a}^{(q)}, \: \forall q =\{0,1,2\} \label{initial_conditions}\\
\hspace{0.5em}&       \mathbf{\underline{p}} \preceq \mathbf{p}_i[k] \preceq \mathbf{\overline{p}},\: \forall k \label{workspace} \\
\hspace{0.5em}&       \left\Vert \dot{\mathbf{p}}_i[k] \right\Vert^2 \leq \overline{v}^2, \:\forall k \label{quad_vel_limits}\\
\hspace{0.5em}&        \underline{f}^2 \leq \left\Vert \ddot{\mathbf{p}}_i[k] + \mathbf{g} \right\Vert^2 \leq \overline{f}^2, \forall k \label{quad_acc_limits} \\
\hspace{0.5em}&          h_{ij}[k] = \left\Vert\boldsymbol\Theta_{ij}^{-1}(\mathbf{p}_i[k] - \boldsymbol\xi_j[k])\right\Vert^2 - 1 \geq 0, \:\forall k,j,
\label{collision_constraint}
\end{align}
\end{subequations}\normalsize
where $k$ is the discrete-time index, $K$ is the planning horizon length, $||\cdot||$ denotes the Euclidean norm, and the superscript $(q)$ denotes the $q$-th time derivative of a variable.

The cost function consists of two terms. The first term is the goal cost that penalizes the deviation of the position of the quadrotor from the specified goal position over the last $\kappa< K$ steps in the prediction horizon; the second term is the smoothness cost that penalizes the $q$-th derivatives of the position trajectory. The constants $w_g$ and $w_s$ are weights trading off the respective cost terms. 

The equality constraints \eqref{initial_conditions} set the initial position of the trajectory and the higher derivatives to be consistent with the current values of the quadrotor.
The inequalities \eqref{workspace}-\eqref{quad_acc_limits} enforce bounds on the position ($\underline{\textbf{p}},\overline{\textbf{p}} $), bounds on the velocity ($-\overline{v}, \overline{v}$), and bounds on the acceleration ($\underline{f}, \overline{f}$). 
The inequalities \eqref{collision_constraint} enforce the collision avoidance requirement with either the $j$-th neighbouring quadrotor or obstacle with position $\boldsymbol\xi_{j}[k]$. The matrix $\boldsymbol\Theta_{ij}$ is a diagonal matrix with $(a, b, c)$ being its element. These scalars $(a, b, c)$ characterize the axis lengths of the ellipsoidal envelopes around the neighbouring quadrotors or the obstacles. The vector $\mathbf{g} = [ 0,\: 0,\: g]^T$ is the gravitational acceleration vector, where $g$ is the acceleration due to gravity.

\textit{Alternative Collision Avoidance Constraint:} The condition in \eqref{collision_constraint} is commonly found in works on quadrotor swarms (e.g., \cite{luis-ral19, luis-ral20, honig2018trajectory, park2022online}). A fundamental problem with the standard collision avoidance constraint~\eqref{collision_constraint} is that these inequalities do not get activated until the planned trajectory intersects with the neighbouring quadrotors or obstacles~\cite{zeng2021safety}. Due to the receding horizon nature of the planning, a quadrotor only tries to avoid collisions with its neighbours or obstacles when it is sufficiently close to them. Increasing the planning horizon can mitigate this issue but at the cost of increased computation time. An alternative approach is to use BF constraints to induce a desired collision avoidance behaviour~\cite{zeng2021safety}: 
\begin{align}
    h_{ij}[k] - h_{ij}[k-1] \geq -\gamma\: h_{ij}[k-1] ,\: \forall k,j 
 \label{cbf_constraints},
\end{align}
\normalsize

where $\gamma \in [0,1]$ is a constant controlling how fast the quadrotor is allowed to approach the constraint boundary given by $h_{ij} = 0$.  Smaller values of $\gamma$ generally result in more gradual and conservative collision avoidance behaviours. With $\gamma = 1$, we recover the original collision-avoidance constraint in~\eqref{collision_constraint}.

\subsection{Trajectory Parameterization}
We parameterize the $x$-, $y$-, and $z$-position trajectories for each quadrotor as Bernstein polynomials of degree~$n$. For instance, the $x$-position trajectory for the $i$-th quadrotor is
\begin{align}
    \begin{bmatrix}x_i[0]&x_i[1]&\ldots&x_i[K-1]
    \end{bmatrix}^T = \mathbf{W}\mathbf{c}_{i,x},
    \label{param}
\end{align}
\normalsize
where $\mathbf{W}\in \mathbb{R}^{K\times(n+1)}$ is the Bernstein basis matrix and $\mathbf{c}_{i,x}$ are the coefficients associated with it. The $k$-th row and $m$-th column element of $\mathbf{W}$ is $[\mathbf{W}]_{km} = {n \choose m} (1-t/(K-1)\delta t)^{n-m}(t/(K-1)\delta t)^{m}$, where $\delta t$ is the discrete-time step size, and $t=k\delta t$ is the continuous-time variable. The higher derivatives of the position trajectory have the general form  $\mathbf{W}^{(q)}\mathbf{c}_{i,x}$, where $\mathbf{W}^{(q)}$ is the $q$-th derivative of the Bernstein basis matrix. The position trajectories for the $y$- and $z$-directions are defined in a similar way.

\subsection{Challenges in Solving the Optimization Problem}
\label{gaps_existing}
 The optimization problem in \eqref{cost}-\eqref{collision_constraint} is a non-convex quadratically constrained quadratic program (QCQP). Existing works (e.g., \cite{augugliaro2012generation,luis-ral19,luis-ral20}) achieve a more favourable QP form by deriving the convex approximation of \eqref{quad_vel_limits}-\eqref{collision_constraint}: the velocity and acceleration bounds are split into axis-wise affine bounds, and the non-convex collision avoidance constraints are approximated as affine constraints through linearization along a trajectory. These approximations can lead to a substantial loss of the feasible space (see Fig.~3 in \cite{mueller2013model}, Fig.~2 in \cite{morgan2016swarm}).

Achieving a QP structure with the BF constraints~\eqref{cbf_constraints} is even more challenging. To see this, we rewrite \eqref{cbf_constraints} as

\begin{align}
    -h_{ij}[k] + (1-\gamma) h_{ij}[k-1] \leq 0,\: \forall k,j.
    \label{bf_convex_discussion}
\end{align}\normalsize
The constraint is non-convex, and the non-convexity comes from the first term $-h_{ij}[k]$. If we linearize the first term, we get an exact but a more conservative convex substitute for the BF constraint. However, the resulting constraint still remains quadratic in the decision variables due to the presence of $h_{ij}[k-1]$. Linearizing the complete right-hand side of~\eqref{bf_convex_discussion} will lead to a QP form.  However, satisfaction of a completely linearized version may not imply satisfaction of \eqref{bf_convex_discussion} \cite{boyd_ccp}.

\section{The Alternating Minimization Algorithm} 
\label{section:sim}
This section presents our main algorithmic results: an AM-based linearization-free trajectory optimizer for solving the motion planning problem introduced in Sec.~\ref{subsec:opt_prob_formulation}. We first present the solution for the case with standard collision constraints \eqref{collision_constraint}. We then show how it naturally extends to the BF constraints \eqref{cbf_constraints}.

\subsection{Constraints Reformulation}
In our proposed approach, one key ingredient that enables us to bring the QCQP problem to a QP form is a polar reformulation of the quadratic constraints \eqref{quad_vel_limits}-\eqref{collision_constraint}. Here we present a general form of the polar representation presented in \cite{adajania2021embedded} for quadratic constraints. 

Consider an inequality of the form $||\mathbf{M}(\mathbf{v}- \mathbf{v}_0)||^2 \le \eta^2$ (or $||\mathbf{M}(\mathbf{v}- \mathbf{v}_0)||^2 \ge \eta^2$) with $\mathbf{M}$ being a diagonal matrix with positive entries. The inequality constraint can be equivalently written in a polar form as follows: $\mathbf{f} = \mathbf{M}(\mathbf{v}- \mathbf{v}_0) - d\:\boldsymbol{\omega}(\alpha, \beta) = 0$ with $d \le \eta$ (or $d \ge \eta$). Here, $\polardir(\alpha,\beta) = [\cos\alpha\sin\beta,\: \sin\alpha\sin\beta,\: \cos\beta]^T$ is a unit direction vector pointing from $\mathbf{v}_0$ to $\mathbf{v}$ with $\alpha$ being the azimuthal angle and $\beta$ being the polar angle, and the scalar $d$ is the magnitude of the vector $\mathbf{M}(\mathbf{v}- \mathbf{v}_0)$.

Using the polar reparametrization, we can write the quadratic constraints \eqref{quad_vel_limits}, \eqref{quad_acc_limits}, and \eqref{collision_constraint} as the following constraint sets:

\begin{align}
 \mathcal{C}_{i,v}[k] &= \{\dot{\mathbf{p}}_i[k]\in\mathbb{R}^3 \:|\: \mathbf{f}_{i,v}[k]=0,  d_{i,v}[k] \leq \overline{v} \label{d_vel}\},\:\forall k,\\
 \mathcal{C}_{i,a}[k] &= \{\ddot{\mathbf{p}}_i[k]\in\mathbb{R}^3 \:|\: \mathbf{f}_{i,a}[k]=0,  \underline{f} \leq d_{i,a}[k] \leq \overline{f} \label{d_acc}\},\:\forall k,\\
 \mathcal{C}_{ij,c}[k] &= \{\mathbf{p}_i[k] \in\mathbb{R}^3\:|\: \mathbf{f}_{ij,c}[k]=0, d_{ij,c}[k] \geq 1 \label{d_coll}\},\:  \forall k,j,
\end{align}\normalsize
where the functions $\mathbf{f}_{ij,c}$, $\mathbf{f}_{i,v}$, and $\mathbf{f}_{i,a}$ are

\begin{align}
\mathbf{f}_{i,v}[k] &= \dot{\mathbf{p}}_i[k] - d_{i,v}[k]\:\polardir(\alpha_{i,v}[k], \beta_{i,v}[k]),\notag\\
\mathbf{f}_{i,a}[k] &= \ddot{\mathbf{p}}_i[k] + \mathbf{g} - d_{i,a}[k]\:\polardir(\alpha_{i,a}[k], \beta_{i,a}[k]),\notag\\
\mathbf{f}_{ij,c}[k] &=\boldsymbol{\Theta}^{-1}_{ij} (\mathbf{p}_i[k] - \boldsymbol\xi_j[k]) - d_{ij,c}[k] \polardir(\alpha_{ij,c}[k], \beta_{ij,c}[k]).\notag
\end{align}\normalsize 
Note that $(\alpha_{\cdot,\cdot}, \beta_{\cdot,\cdot}, d_{\cdot,\cdot})$ are the parameters of the polar form representations of the constraints and will be computed by our optimizer together with the trajectory.

\subsection{Reformulated Problem}
Before deriving the final form of our reformulated problem, we rewrite the polar form constraints derived in the previous subsection in a compact matrix form. Given the parametrization in \eqref{param}, the equality part of constraints \eqref{d_vel}, \eqref{d_acc}, \eqref{d_coll} can be represented as 

%
\begin{align}
\overbrace{
\begin{bmatrix}
    \widetilde{\mathbf{A}} &&\\
    &\widetilde{\mathbf{A}} &\\
    &&\widetilde{\mathbf{A}}
\end{bmatrix}}^{\mathbf{A}}
\overbrace{\begin{bmatrix}
    \mathbf{c}_{i,x}\\
    \mathbf{c}_{i,y}\\
    \mathbf{c}_{i,z}
\end{bmatrix}}^{\boldsymbol\zeta_{i,1}} =
\overbrace{
\begin{bmatrix}
        \mathbf{d}_{i,v}\cos\boldsymbol{{\alpha}}_{i,v}\sin\boldsymbol{{\beta}}_{i,v}\\ 
        \mathbf{d}_{i,a}\cos\boldsymbol{{\alpha}}_{i,a}\sin\boldsymbol{{\beta}}_{i,a}\\
        \boldsymbol{\xi}_x + a\mathbf{d}_{i,c}\cos\boldsymbol{{\alpha}}_{i,c}\sin\boldsymbol{{\beta}}_{i,c}\\
        \hline
        \mathbf{d}_{i,v}\sin\boldsymbol{{\alpha}}_{i,v}\sin\boldsymbol{{\beta}}_{i,v}\\
        \mathbf{d}_{i,a}\sin\boldsymbol{{\alpha}}_{i,a}\sin\boldsymbol{{\beta}}_{i,a}\\
        \boldsymbol{\xi}_y + b\mathbf{d}_{i,c}\sin\boldsymbol{{\alpha}}_{i,c}\sin\boldsymbol{{\beta}}_{i,c}\\
        \hline
        \mathbf{d}_{i,v}\cos\boldsymbol{{\beta}}_{i,v}\\
        -\widetilde{\mathbf{g}} + \mathbf{d}_{i,a}\cos\boldsymbol{{\beta}}_{i,a}\\
        \boldsymbol{\xi}_z + c\mathbf{d}_{i,c}\cos\boldsymbol{{\beta}}_{i,c}
\end{bmatrix}.
}^{\mathbf{b}}
\end{align} \normalsize

Here, $\widetilde{\mathbf{A}} =  \begin{bmatrix}
        \dot{\mathbf{W}}^T & \ddot{\mathbf{W}}^T & \mathbf{F}_{c}^T  
    \end{bmatrix}^T$, the matrix $\mathbf{F}_{c}$ is constructed by vertically stacking the matrix $\mathbf{W}$ as many times as the number of neighbouring quadrotors and obstacles present in the environment. The vectors $(\boldsymbol{\xi}_x,\boldsymbol{\xi}_y,\boldsymbol{\xi}_z)$ are formed by vertically stacking the corresponding variables $({\xi}_{j,x}[k],{\xi}_{j,y}[k],{\xi}_{j,z}[k])$ at different time steps of the prediction horizon and for all the neighbouring quadrotors and obstacles. In a similar fashion, $(\boldsymbol{\alpha}_{\cdot,\cdot}, \boldsymbol{\beta}_{\cdot,\cdot}, \mathbf{d}_{\cdot,\cdot})$ are formed by vertically stacking $({\alpha}_{\cdot,\cdot}[k], {\beta}_{\cdot,\cdot}[k], {d}_{\cdot,\cdot}[k])$. The vector  $\widetilde{\mathbf{g}}$ is formed by vertically stacking  $g$ as many times as the length of prediction horizon.

Using the derivations above, we now write the reformulated optimization problem as


\begin{subequations}
    \begin{align}
    \min_{\boldsymbol{\zeta}_{i,1}, \boldsymbol{\zeta}_{i,2}, \boldsymbol{\zeta}_{i,3}} \hspace{0.5em}&\frac{1}{2} \boldsymbol{\zeta}_{i,1}^T \mathbf{Q} \boldsymbol{\zeta}_{i,1} + \mathbf{q}^T\boldsymbol{\zeta}_{i,1} \label{cost_matrix}\\
   \text{s.t.}\hspace{0.5em}& \mathbf{A} \boldsymbol{\zeta}_{i,1} = \mathbf{b}(\boldsymbol{\zeta}_{i,2}, \boldsymbol{\zeta}_{i,3}) \label{eq_matrix}\\  \hspace{0.5em}&\mathbf{G} \boldsymbol{\zeta}_{i,1} \preceq \mathbf{h} \label{ineq_matrix}\\
   \hspace{0.5em}& \boldsymbol{\zeta}_{i,1} \in \mathcal{C}_{{\zeta}_{i,1}}, \boldsymbol{\zeta}_{i,3} \in \mathcal{C}_{{\zeta}_{i,3}},\label{initial_condi_matrix}
\end{align}
\end{subequations} \normalsize
where $\boldsymbol{\zeta}_{i,1} = [\mathbf{c}_{i,x}^T, \:\mathbf{c}_{i,y}^T,\:\mathbf{c}_{i,z}^T]^T$, $\boldsymbol{\zeta}_{i,2} =[\boldsymbol{\alpha}_{i,c}^T,\: \boldsymbol{\alpha}_{i,a}^T,\: \boldsymbol{\alpha}_{i,v}^T,\:  \boldsymbol{\beta}_{i,c}^T$, $  \boldsymbol{\beta}_{i,a}^T,\: \boldsymbol{\beta}_{i,v}^T]^T$, and $\boldsymbol{\zeta}_{i,3} = [\mathbf{d}_{i,c},\: \mathbf{d}_{i,a},\: \mathbf{d}_{i,v}]^T$ are the variables to be optimized. The matrix $\mathbf{Q}$ and vector $\mathbf{q}$ are formed from the objective function \eqref{cost}. The matrix $\mathbf{G}$ and vector $\mathbf{h}$ in the inequality constraint \eqref{ineq_matrix} stem from the positional bounds \eqref{workspace}. 
 
 
 The set $\mathcal{C}_{\boldsymbol{\zeta}_{i,1}} \mathord{=} \{\boldsymbol{\zeta}_{i,1}\in\mathbb{R}^{3n}\:|\:\mathbf{C}\boldsymbol{\zeta}_{i,1}=\mathbf{e}\}$ encodes the initial conditions \eqref{initial_conditions}. Here, the matrix $\mathbf{C} = [
     \mathbf{W}_0^T,\: \dot{\mathbf{W}}_0^T,\: \ddot{\mathbf{W}}_0^T  
 ]^T$, and the subscript $0$ represents the $0$-$th$ row of the respective matrices. The vector $\mathbf{e}=[
     \mathbf{p}_{i,a}^T,\: \dot{\mathbf{p}}_{i,a}^T,\: \ddot{\mathbf{p}}_{i,a}^T 
 ]^T$ contains the current position, velocity and acceleration values. The set $\mathcal{C}_{\boldsymbol{\zeta}_{i,3}}$ consists of the conditions on each of the variables $(\mathbf{d}_{ij,c}, \mathbf{d}_{i,v}, \mathbf{d}_{i,a})$ derived from the polar reformulation.
 
 
\subsection{Relaxation and Solution by AM}
\label{subsection::am}
The optimization \eqref{cost_matrix}-\eqref{initial_condi_matrix} has some hidden convex structures which makes it suitable for AM-based approaches. To exploit these structures, we first relax the non-convex equality \eqref{eq_matrix} and affine \eqref{ineq_matrix} constraints as penalties in the following form:


\begin{align}
    \min_{\boldsymbol{\zeta}_{i,1} \in \mathcal{C}_{{\zeta}_{i,1}}, \boldsymbol{\zeta}_{i,3} \in \mathcal{C}_{{\zeta}_{i,3}}} \frac{1}{2} \boldsymbol{\zeta}_{i,1}^T\mathbf{Q}\boldsymbol{\zeta}_{i,1} + \mathbf{q}^T\boldsymbol{\zeta}_{i,1} - \langle \boldsymbol\lambda_i, \boldsymbol{\zeta}_{i,1}\rangle\notag \\+ \frac{\rho}{2}\left \Vert \mathbf{A} \boldsymbol{\zeta}_{i,1} - \mathbf{b}(\boldsymbol{\zeta}_{i,2}, \boldsymbol{\zeta}_{i,3}) \right \Vert^2 + \frac{\rho}{2}\left \Vert \mathbf{G} \boldsymbol{\zeta}_{i,1} - \mathbf{h} + {\mathbf{s}_i} \right \Vert^2. \label{augmented_problem}
\end{align} \normalsize
The parameter $\rho$ trades-off satisfaction of constraint residual with the minimization of primary cost function. The slack variable $\textbf{s}_i\geq \mathbf{0}$ is unknown, and we discuss shortly how these are obtained within an AM setup. The vector $\boldsymbol{\lambda}_i$ is called the Lagrange multiplier and is crucial for driving the constraint residuals to zero \cite{admm_neural}. 


Algorithm \ref{algo_1} summarizes the AM steps for minimizing \eqref{augmented_problem}, wherein the superscript $l$ in $^{l}{(.)}$ represents the value of $(.)$ at $l$-th iteration of the algorithm. At each step of the AM, only one of $\boldsymbol\zeta_{i,1}, \boldsymbol\zeta_{i,2}, \boldsymbol\zeta_{i,3} $ are optimized while the rest are held fixed at values obtained in the previous update. Each step in Algorithm \ref{algo_1} is either a convex QP or has a closed-form solution. We discuss these observations below in detail.

\textit{Step~\eqref{solve_xyz}}: We solve for $\boldsymbol\zeta_{i,1}$ while keeping the other variables constant. We see that the problem is an equality-constrained convex QP whose solution boils down to solving a set of linear equations:

    %
    \begin{align}
    \begin{bmatrix}
    \check{\mathbf{A}} & \mathbf{C}^{T}\\
    \mathbf{C} & \mathbf{0} 
    \end{bmatrix}
    \begin{bmatrix}
        ^{l+1}\boldsymbol{\zeta}_{i,1}\\
        \boldsymbol{\mu}
        \end{bmatrix} =\begin{bmatrix}
        ^l\check{\mathbf{b}}\\
        \mathbf{e}
        \end{bmatrix}, \label{sol_xyz}
     \end{align} \normalsize
where the matrix $\check{\mathbf{A}} \mathord{=} \mathbf{Q}+\rho\mathbf{G}^T\mathbf{G} + \rho\mathbf{A}^T\mathbf{A}$,  the vector $^l\check{\mathbf{b}} \mathord{=} \mathbf{q}+ \rho\mathbf{G}^T(\mathbf{h}-^{l}\mathbf{s}_i) + \rho\mathbf{A}^T {^l}\mathbf{b}-{^l}\boldsymbol{\lambda}_i$, and $\boldsymbol\mu$ are the dual variables associated with the equality constraints.

\textit{Step~\eqref{solve_angles}}: We now solve for $\boldsymbol\zeta_{i,2}$. As an example, the optimization problem for the variables $(\boldsymbol\alpha_{i,c}, \boldsymbol\beta_{i,c})$ is
    \begin{align}
        {^{l+1}}\boldsymbol{\alpha}_{i,c}, {^{l+1}}\boldsymbol{\beta}_{i,c} \mathord{=} \arg \min_{\alpha_{i,c}, \beta_{i,c}} \notag\\ \left\Vert  \mathbf{F}_{c}{^{l+1}}\mathbf{c}_{i,x} - \boldsymbol{\xi}_x - a{^{l}}\mathbf{d}_{i,c}\cos\boldsymbol{{\alpha}}_{i,c}\sin\boldsymbol{{\beta}}_{i,c} \right\Vert^2 \notag\\+ \left\Vert \mathbf{F}_{c}{^{l+1}}\mathbf{c}_{i,y} -
        \boldsymbol{\xi}_y - b{^{l}}\mathbf{d}_{i,c}\sin\boldsymbol{{\alpha}}_{i,c}\sin\boldsymbol{{\beta}}_{i,c} \right\Vert^2 \notag\\ + \left\Vert  \mathbf{F}_{c}{^{l+1}}\mathbf{c}_{i,z}-
        \boldsymbol{\xi}_z - c{^{l}}\mathbf{d}_{i,c}\cos\boldsymbol{{\beta}}_{i,c} \right\Vert^2.
        \label{alpha_beta_coll}
    \end{align} \normalsize
The minimization \eqref{alpha_beta_coll} is simply a projection of $(^{l+1}{\mathbf{c}_{i,x}}, ^{l+1}{\mathbf{c}_{i,y}}, ^{l+1}{\mathbf{c}_{i,z}})$ onto ellipsoids centered at $(\boldsymbol{\xi}_x, \boldsymbol{\xi}_y, \boldsymbol{\xi}_z )$ and has a closed-form solution \cite{adajania2021embedded}. Similarly, we can obtain $(^{l+1}\boldsymbol{\alpha}_{i,v}, ^{l+1}\boldsymbol{\beta}_{i,v})$ and $(^{l+1}\boldsymbol{\alpha}_{i,a}, ^{l+1}\boldsymbol{\beta}_{i,a})$. 

\textit{Step~\eqref{solve_d}}: The optimization over $\textbf{d}_{i, c}$ involves solving the following QP:

    \begin{align}
        {^{l+1}}\mathbf{d}_{i,c} = \arg \min_{\mathbf{d}_{i,c} \geq 1} \notag\\ \left\Vert  \mathbf{F}_{c}{^{l+1}}\mathbf{c}_{i,x} - \boldsymbol{\xi}_x - a\mathbf{d}_{i,c}\cos{^{l+1}}\boldsymbol{{\alpha}}_{i,c}\sin{^{l+1}}\boldsymbol{{\beta}}_{i,c} \right\Vert^2 \notag\\+ \left\Vert \mathbf{F}_{c}{^{l+1}}\mathbf{c}_{i,y} -
        \boldsymbol{\xi}_y - b\mathbf{d}_{i,c}\sin{^{l+1}}\boldsymbol{{\alpha}}_{i,c}\sin{^{l+1}}\boldsymbol{{\beta}}_{i,c} \right\Vert^2 \notag\\ + \left\Vert  \mathbf{F}_{c}{^{l+1}}\mathbf{c}_{i,z}-
        \boldsymbol{\xi}_z - c\mathbf{d}_{i,c}\cos{^{l+1}}\boldsymbol{{\beta}}_{i,c} \right\Vert^2.
        \label{d_coll_sol}
    \end{align}\normalsize
Each element of $\textbf{d}_{i, c}$ is decoupled from each other. Thus \eqref{d_coll_sol} reduces to parallel single variable QPs, each of which can be solved in closed form. We clip the resulting solution to $(1, \infty)$ to satisfy the  lower bound on $\textbf{d}_{i, c}$ (recall the definition of $\mathcal{C}_{{\zeta}_{i,3}}$). We obtain $\mathbf{d}_{i,v}$ and $\mathbf{d}_{i,a}$ in a similar fashion with their respective clipping bounds.
    

\textit{Step~\eqref{update_slack}}: The slack variables $\mathbf{s}_i$ is updated based on \cite{ghadimi2014optimal}.

\textit{Step~\eqref{update_lagrange}}: The Lagrange multipliter $\boldsymbol\lambda_i$ is updated using the approach presented in \cite{admm_neural}.



\begin{algorithm}[!t]
\centering
 \small
 \caption{The AM algorithm used by quadrotor $i$ at each planning step}\label{algo_1}
    \begin{algorithmic}[1]   
    \State Initialize ${^{l}}\boldsymbol{\zeta}_{i,2}=\mathbf{0}, {^{l}}\boldsymbol{\zeta}_{i,3}=\mathbf{0}, {^{l}}\boldsymbol{\lambda}_i=\mathbf{0}, {^{l}}\mathbf{s}_{i}=\mathbf{0}$ at $l = 0$
    \While {$l\leq \text{maxiter}$ or \text{residuals} $\geq$ \text{thresold}}
    \State \begin{align}
        {^{l+1}}\boldsymbol{\zeta}_{i,1} &=  \text{arg} \min_{\boldsymbol{\zeta}_{i,1} \in \mathcal{C}_{\boldsymbol{\zeta}_{i,1}}} \frac{1}{2} \boldsymbol{\zeta}_{i,1}^T \mathbf{Q}\boldsymbol{\zeta}_{i,1} + \mathbf{q}^T\boldsymbol{\zeta}_{i,1} - \langle {^{l}}\boldsymbol\lambda, \boldsymbol{\zeta}_{i,1}\rangle\notag\\         +& \frac{\rho}{2}\left \Vert \mathbf{A} \boldsymbol{\zeta}_{i,1} - \mathbf{b}({^{l}}\boldsymbol{\zeta}_{i,2}, {^{l}}\boldsymbol{\zeta}_{i,3}) \right \Vert^2 + \frac{\rho}{2}\left \Vert \mathbf{G} \boldsymbol{\zeta}_{i,1} - \mathbf{h} + {^{l}}\mathbf{s}_{i} \right \Vert^2 \label{solve_xyz} \tag{S1}
        \end{align}\vspace{-0.5em}
        \begin{align}
        {^{l+1}\boldsymbol{\zeta}_{i,2}} =& \text{arg} \min_{\boldsymbol{\zeta}_{i,2}} \frac{\rho}{2}\left \Vert \mathbf{A} {^{l+1}}\boldsymbol{\zeta}_{i,1} - \mathbf{b}(\boldsymbol{\zeta}_{i,2}, {^{l}}\boldsymbol{\zeta}_{i,3}) \right \Vert^2 \label{solve_angles}\tag{S2}\\
        {^{l+1}\boldsymbol{\zeta}_{i,3}} =& \text{arg} \min_{\boldsymbol{\zeta}_{i,3} \in \mathcal{C}_{\boldsymbol{\zeta}_{i,3}}} \frac{\rho}{2}\left \Vert \mathbf{A} {^{l+1}}\boldsymbol{\zeta}_{i,1} - \mathbf{b}({^{l+1}}\boldsymbol{\zeta}_{i,2}, \boldsymbol{\zeta}_{i,3}) \right \Vert^2 \label{solve_d} \tag{S3}\\
        {^{l+1}}\mathbf{s}_{i} =& \text{max}\left(0, -\mathbf{G} {^{l+1}}\boldsymbol{\zeta}_{i,1} - \mathbf{h}\right) \label{update_slack} \tag{S4}\\
        {^{l+1}}\boldsymbol{\lambda}_i =& {^{l}}\boldsymbol{\lambda}_i - \frac{\rho}{2} \mathbf{A}^T\left( \mathbf{A} {^{l+1}}\boldsymbol{\zeta}_{i,1} - \mathbf{b}({^{l+1}}\boldsymbol{\zeta}_{i,2}, {^{l+1}}\boldsymbol{\zeta}_{i,3})\right) \notag\\-& \frac{\rho}{2}\mathbf{G}^T\left(\mathbf{G} {^{l+1}}\boldsymbol{\zeta}_{i,1} - \mathbf{h} + {^{l+1}}\mathbf{s}_i\right) \label{update_lagrange}\tag{S5}
    \end{align} 
    \EndWhile
    \normalsize
    \end{algorithmic}
\end{algorithm}


\subsection{Incorporating BF Constraints}
In \eqref{d_coll}, $d_{i, c}[k] \mathord{=} 1$ corresponds to the boundary of the feasible set of the collision avoidance constraints \eqref{collision_constraint}. In other words, $d_{ij, c}[k] \mathord{=} 1$ ensures $h_{ij}[k] \mathord{=} 0$. Along similar lines, $d_{ij, c}[k] < 1$ would correspond to the interior of the set. With this insight, we can define the polar reformulation of BF constraints in the following form:
\begin{align}
&\mathcal{C}_{ij,bf}[k] = \{\mathbf{p}_i[k] \:|\: \mathbf{f}_{ij,c}[k]=0,\nonumber\\&\hspace{2em} d_{ij,c}[k] \geq 1 + (1-\gamma) (d_{ij,c}[k-1] - 1) \label{d_coll_cbf}\},\:  \forall k,j,
\end{align}
By comparing \eqref{d_coll} and \eqref{d_coll_cbf}, we can see that the constraints differ only in the feasible region definition of $d_{ij, c}[k]$. When $\gamma \mathord{=} 1$, the constraints are equivalent. 

We integrate \eqref{d_coll_cbf} into Algorithm \ref{algo_1} through a minor modification in \textit{Step~\eqref{solve_d}}, specifically QP \eqref{d_coll_sol}. Let $^{l}d_{ij,c}[k]$ be the value of $d_{ij,c}[k]$ obtained at the $l$-th iteration of Algorithm \ref{algo_1}. We can use this value to approximate the feasible region of $d_{ij, c}[k]$ for BF constraints at the $(l+1)$-th iteration as
\begin{align}
    d_{ij,c}[k] \geq 1 + (1-\gamma) ({^l}d_{ij,c}[k-1] - 1).
    \label{d_coll_cbf_approx}
\end{align}
The right-hand side of \eqref{d_coll_cbf_approx} is constant, and thus the feasible region of $d_{ij,c}[k] $ for BF constraints is approximated through a simple lower bound. We can now just solve the QP \eqref{d_coll_sol} and clip the obtained value to this lower bound to solve for the optimal $d_{ij,c}[k] $ at iteration $(l+1)$.

\section{Simulation Results}
\label{section:sim}
This section provides a simulation analysis and comparison of our approach against the state-of-the-art baselines \cite{luis-ral19,dmpc_epfl,acado}. We denote our approach as ``\text{Ours (Quadratic)}'', and the ``\text{Quadratic}'' in the parenthesis refers to quadratic kinematic constraints. The proposed approach and the baselines are implemented in C++. The codes are available here \cite{github_repo_amswarm}. All the simulations were executed on a PC with Intel Xeon CPU with 8 cores and 16 GB of RAM, running at 3 GHz.

The prediction horizon length is set to $K \mathord{=} 30$ with a discretization of $0.1s$. The Bernstein polynomials used to parameterize the position trajectories have a degree of $n\mathord{=}10$. In the trajectory optimization problem \eqref{cost}-\eqref{collision_constraint}, we set $w_g \mathord{=} 7000$, $w_s\mathord{=}100$, $\kappa\mathord{=}5$ and penalize the acceleration $(q\mathord{=}2)$ trajectory in the cost. In the constraints, we set $\overline{v}\mathord{=}1.73ms^{-1}$, $\underline{f} \mathord{=} 0.3g$ and $\overline{f}\mathord{=}1.5g$. For collision avoidance with neighbouring quadrotors, we set ${\boldsymbol\Theta}_{ij} \mathord{=} \diag(0.17m,0.17m,0.45m)$, but a collision is declared with ${\boldsymbol\Theta}_{coll} \mathord{=} \diag(0.13m,0.13m,0.40m)$. A quadrotor $j$ is considered to be a potential conflict for quadrotor $i$ if at any prediction step of the horizon, $\left\Vert(\boldsymbol\Theta_{ij}+\boldsymbol\Theta_{p})^{-1}(\mathbf{p}_i[k] - \boldsymbol\xi_j[k])\right\Vert^2 \mathord{\geq} 1$ holds, where $\boldsymbol\Theta_p\mathord{=}\diag(0.2m,0.2m,0.2m)$.
Similarly, we choose appropriate parameters for the collision constraints with obstacles. In Algorithm \ref{algo_1}, we set maxiter$\mathord{=}2000$, thresold$\mathord{=}0.01$, and the penalty parameter $\rho\mathord{=}min(1.3^l, 5e10^5)$, where $l$ is the iteration count of the algorithm. 


\raggedbottom
\subsection{Distributed Swarm Baselines}
We compare our proposed approach \text{Ours (Quadratic)} with $\gamma\mathord{=}1$ with three different distributed swarm baselines: 
\begin{enumerate}
    \item \text{SCP (On-demand)} \cite{luis-ral20}: This approach relies on linearization of collision avoidance constraints and axis-wise decoupling of the kinematic bounds. It uses a so-called On-demand strategy where it tries to resolve only the first predicted collision. \item \text{SCP (Continuous)} \cite{dmpc_epfl}: This approach is similar to \text{SCP (On-demand)} but it adds collision constraints over the entire prediction horizon. \item \text{Ours (Axiswise)}: This approach is the same as \text{Ours (Quadratic)} in the sense that it does not rely on linearizing collision constraints but has axis-wise kinematic bounds. The maximum and minimum values for the axis-wise velocity bounds are $\pm{\overline{v}}/{\sqrt3}$, and the acceleration limits can be computed such that it satisfies extreme cases (see (13) and (14) of \cite{augugliaro2012generation}).
\end{enumerate}

We consider a cluttered environment with $16$ cylindrical static obstacles in a volume of $4m\times4m\times2m$ and vary swarms from $10$ to $50$. We tested the approaches in $100$ configurations for each swarm size with randomized start-goal and obstacle positions.  A trial is successful if all the quadrotors reach their assigned goal positions without collisions and under a time limit of $20s$. 

\subsection{Comparative  Analysis}

\begin{figure}[!t]
	\centering
	\begin{minipage}{.325\columnwidth}
		\centering
		\includegraphics[width=\textwidth]{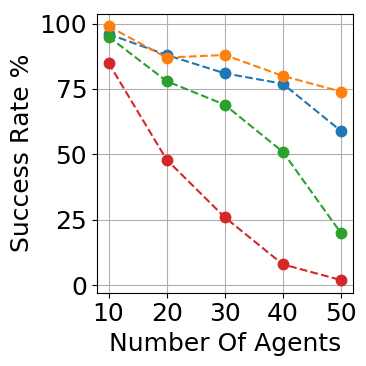}
	\end{minipage}~
	\begin{minipage}{.325\columnwidth}
		\centering
		\includegraphics[width=\textwidth]{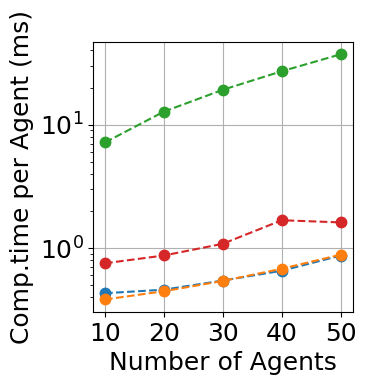}
	\end{minipage}~
	\begin{minipage}{.325\columnwidth}
		\centering
		\includegraphics[width=\textwidth]{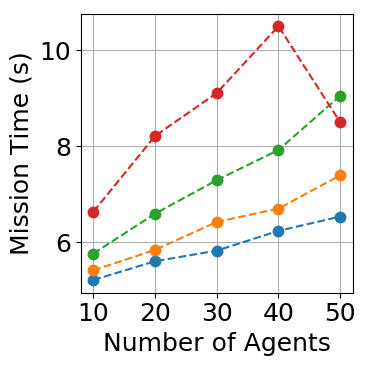}
	\end{minipage}
     \centering
	   \includegraphics[width=\columnwidth]{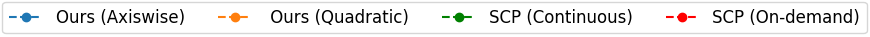}
	\caption{Average performance comparison of approaches in point-to-point transition setting with an increasing number of swarm sizes in a fixed volume of $32m^3$ and with $16$ static obstacles. 100 configurations were run for each swarm size.}
 	\label{comparison_fig_1}
\end{figure}

\textit{Success Rate:} The first plot in Fig. \ref{comparison_fig_1} summarizes the improvement achieved by \text{Ours (Quadratic)} and \text{Ours (Axiswise)} over \text{SCP (On-demand)}, and \text{SCP (Continuous)} in the success-rate metric. For swarm size up to 30, our approaches provide around $11\%$-$62\%$ improvement over  \text{SCP (On-demand)} and $1\%$-$19\%$ over \text{SCP (Continuous)}. As swarm size increase to 50, the performance gap between our approaches and the \text{SCP (On-demand)} and \text{SCP (Continuous)} swell to $39\%$-$72\%$. The explanation is that the SCP baselines replace the non-convex collision constraints as hyperplane constraints, a conservative approximation of free space. Furthermore, SCP (On-demand) adds only a handful of collision constraints that result in unsafe separation and, ultimately, collisions in highly cluttered settings. For swarm size $50$, we also see that Ours (Quadratic) has $15\%$ more success rate than Ours (Axiswise) as the former has larger access to acceleration and velocity bounds.

\textit{Computation Time per Agent}: The middle plot in Fig. \ref{comparison_fig_1} shows the average computation time per agent for the different swarm sizes. Both of our approaches have similar computation time per agent and are substantially faster than \text{SCP (On-demand)} and \text{SCP (Continuous)}.  For the swarm size of $10$ case, we see that our approaches are $1.7$ times faster than SCP (On-demand), and $18.7$ times faster than SCP (Continuous).  With a swarm size of $50$, our approaches are still $1.7$ times faster than SCP (On-demand) but are $42$ times faster than SCP (Continuous).  This excellent performance of our approaches can be attributed to the fact that AM optimizer of Algorithm~\ref{algo_1} requires us to solve only one equality-constrained QP per iteration.  On the contrary, the \text{SCP (Continuous)} solves constrained QP with a large number of inequality constraints.  Moreover, it has to incorporate one slack variable per collision constraint to prevent potential in-feasibility in the optimization problem.  The strategy of incorporating only the most imminent collision in \text{SCP (On-demand)} offers improvement in computation time per agent but at the expense of success rate as described previously.

\textit{Mission Time:} The rightmost plot in Fig. \ref{comparison_fig_1} presents the average time taken by the swarm to reach their desired goal positions. We see that SCP (On-demand) performs the worst as the on-demand strategy leads to agents being closer to each other and obstacles, thus, slowing down the progress toward their goals. SCP (Continuous) performs relatively better than SCP (On-demand), but our approaches have the lowest mission times. For swarm size $40$, we see that Ours (Quadratic) on average shows $36\%$ and $15\%$ reduction in mission time than SCP (On-demand) and SCP (Continuous), respectively. Interestingly, the trends show that Ours (Axiswise), on average, completed the task $0.5s$-$0.9s$ faster than Ours (Quadratic).   

\subsection{Trade-off Between Performance and Safety}
\label{subsection:trade_off}
\begin{figure}[!t]
	\centering
	\begin{minipage}{.325\columnwidth}
		\centering
		{\label{mission_2}\includegraphics[width=\textwidth]{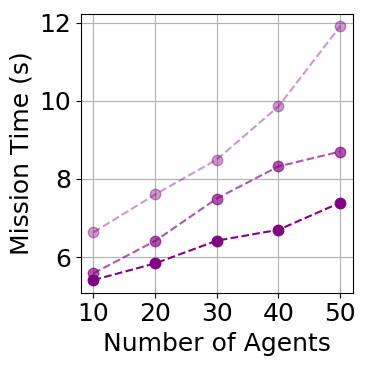}}
	\end{minipage}~
	\begin{minipage}{.325\columnwidth}
		\centering
		{\label{inter_dist_2}\includegraphics[width=\textwidth]{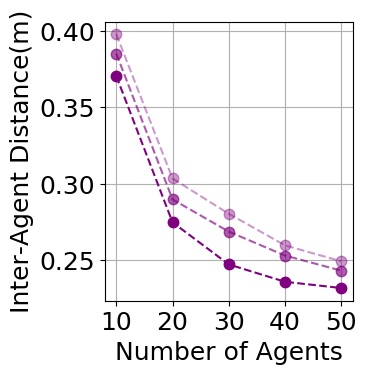}}
	\end{minipage}~
	\begin{minipage}{.325\columnwidth}
		\centering
		{\label{obs_dist_2}\includegraphics[width=\textwidth]{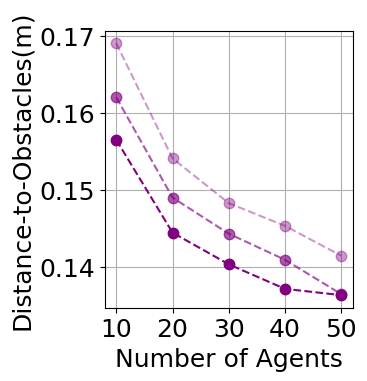}}
	\end{minipage}\\
	\subfigure{\label{legend_2}\includegraphics[width=\columnwidth]{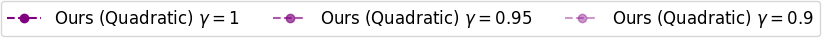}}
	\caption{Average performance comparison of our approach in point-to-point transition setting with different values of $\gamma$ in the barrier function constraints. The environment is of $32m^3$ in volume and has $16$ static obstacles. For each swarm size, $100$ configurations were executed.}
	\label{comparison_2}
\end{figure}

Figure \ref{comparison_2} showcases the performance comparison of \text{Ours (Quadratic)} with three different values of $\gamma$. We ran the same $100$ configurations and recorded the smallest inter-agent distance and distance-to-obstacles observed at each time step. With $\gamma\mathord{=}0.95$, we see $3.98\%$-$8.69\%$ improvement in inter-agent distances over $\gamma\mathord{=}1$. The improvements increase to $7.11\%$-$12.35\%$ with a more conservative $\gamma\mathord{=}0.9$. We see a similar trend in distance-to-obstacles. This improvement in clearance comes at the expense of increased mission time (see first plot in Fig. \ref{comparison_2}). We also observed that the computation time  per agent increased with decreasing $\gamma$. For swarm size $50$, the average computation time per agent is $1.61ms$, $82\%$ increase over $\gamma\mathord{=}1$. Nevertheless, our approach is still real-time.

\subsection{Off-the-shelf Solver Baseline}
\begin{figure}[!t]
		\centering
		{\label{acado_runtime}\includegraphics[width=0.325\columnwidth]{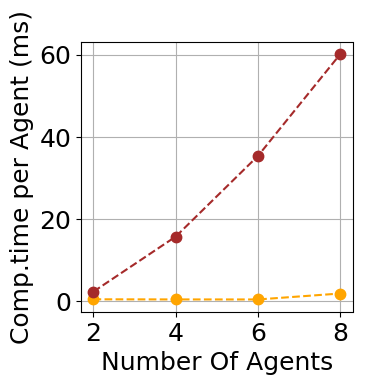}}~
		{\label{acado_mission_time}\includegraphics[width=0.325\columnwidth]{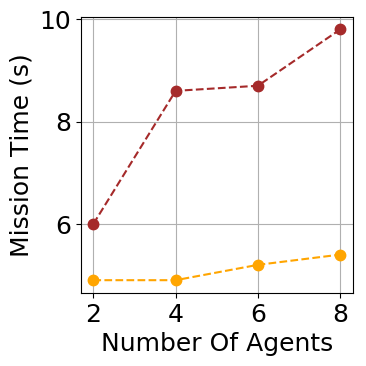}}
	\includegraphics[width=0.55\columnwidth]{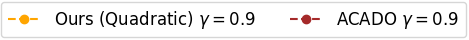}
	\caption{Performance comparison of our approach and ACADO with BF constraints ($\gamma=0.9$) in  antipodal position exchange  scenario with no static obstacles. ACADO could accomodate only a maximum of 8 agents.}
	\label{comparison_3}
\end{figure}

We now compare \text{Ours (Quadratic)} against the optimal control solver ACADO \cite{acado}. ACADO is provided with the original BF constraints \eqref{cbf_constraints}, while \text{Ours (Quadratic)} uses the reformulation presented in \eqref{d_coll_cbf}. We ran an antipodal position exchange with a swarm size of $2$, $4$, $6$, and $8$ with no static obstacles. Figure \ref{comparison_3} presents the observed metrics. We see that the computation time per agent scaling for \text{Ours (Quadratic)} is linear, while for ACADO, it increases quadratically with the number of agents. Furthermore, \text{Ours (Quadratic)} can complete the task faster than ACADO. 



\section{Experimental Evaluation}
We tested our approach on our Crazyflie 2.0 swarm testbed. The quadrotors' trajectories are computed on a single computer, and we send position and velocity trajectories to the underlying lower controller based on \cite{mellinger2011minimum}. More details about the testbed can be found here \cite{github_repo_amswarm}.
 
 A video summarizing the experimental results can be found here: \hyperlink{http://tiny.cc/AMSwarmVideo}{\url{http://tiny.cc/AMSwarmVideo}}. The algorithm is tested on various challenging scenarios. First, a 12 quadrotor swarm performs a head-on transition where a single quadrotor is in conflict with other 11 quadrotors at a time in an obstacle-free environment. Second, we repeat the same transition but with 6 static obstacles in the environment. We see that our approach navigates the quadrotors to their desired goals in an agile and smooth manner. Third, we qualitatively show a 12 drone random transition in an obstacle-free setting with two values of $\gamma$. We see quadrotors with conservative $\gamma$ show a safe, evasive behaviour. Lastly, a formation of 8 quadrotors performs a transition in the presence of an unpredictable human. With the help of  BF constraints, each quadrotor is able to navigate around the human safely.

\section{Conclusion} 
We presented a novel contribution toward making quadrotor swarm navigation more reliable and scalable. We showed how to formulate the original problem with quadratic collision and kinematic constraints as a QP without relying on conservative approximations. Furthermore, our approach can naturally handle more sophisticated collision avoidance constraints based on discrete-time BFs. In simulation, our optimizer significantly outperformed SCP-based approaches that have been common in recent works. Similarly, our approach also proved to be more computationally efficient than the state-of-the-art optimal control solver ACADO when considering BF constraints. In experiments, we showed the efficacy of the proposed algorithm in challenging scenarios in both obstacle-free and cluttered environments.

\bibliography{main}
\bibliographystyle{IEEEtran}

\end{document}